# Identification of arabic word from bilingual text using character features
Case of structural features

Sofiene Haboubi, Samia Maddouri, Hamid Amiri
System and Signal Processing Laboratory
National Engineering School of Tunis (ENIT)
BP 37, Belvedere 1002, Tunis, Tunisia.
sofiene.haboubi@istmt.rnu.tn
{samia.maddouri; hamid.amiri}@enit.rnu.tn

*Abstract*— **The identification of the language of the script is an important stage in the process of recognition of the writing. There are several works in this research area, which treat various languages. Most of the used methods are global or statistical. In this present paper, we study the possibility of using the features of scripts to identify the language. The identification of the language of the script by characteristics returns the identification in the case of multilingual documents less difficult. We present by this work, a study on the possibility of using the structural features to identify the Arabic language from an Arabic / Latin text.**

*Language identification, structural features, Arabic script, Latin script.*

## I. Introduction

The feature extraction is an important step in the process of recognition of handwritten Arabic script. It can generate the primitive descriptive image. The choice of these primitives is crucial for recognition. A compromise must be respected during feature extraction: The feature extractor is to provide primitive uniform for different types of writing, accurately reflecting all the information necessary to the process of recognition. It should also be little greedy execution time in memory [3]. Various types of methods of feature extraction are known in literature [5]. They are based on three main descriptions: the projection of the image [6,13], skeletonization [10, 15] and function contours [14, 17, 18, 20, 21].

The projection method is the simplest and least expensive for small images but it is very sensitive to tilt and it does not accurately describe the image.

The skeletonization is used mainly in order to segment graphemes in writing uniform. The challenge of segmentation of cursive reduces the effectiveness of this description.

The outline, which describes the overall picture, can be easily detected. Subsequent treatments can generate different types of global or analytic description from the outline. This type of description is used in our approach.

The handwriting recognition is a task complicated by the diversity of styles of writing, first and richness of words in different types of information on the other. This is especially true as the Arabic words consist of whole areas.

## II. Characteristics of the Arabic script

We have found it useful to recall briefly the essential characteristics of handwritten Arabic script. This presentation is based on a literature review that we find in [1, 2], ... The Arabic alphabet has 28 letters have different shapes. These forms vary according to their positions in the word, their widths, number and position of the diacritical dots and hamza presence of diacritics.

- Cursivity: The Arabic script is cursive, i.e. the letters are connected. This property is encountered in both printed and handwritten forms.

- Character position: The shapes of Arabic letters depending on their position in the word initial position, medial, final or isolated, which increases the total number of forms of different characters over 100.

- Character Width: Unlike printed Latin characters, hunting Arabic characters printed and handwritten result is variable although they have neither upper nor lower case.

- Diacritics: More than half of Arab characters have diacritical points in their forms. These points are the number one, two or three and can be top or bottom of bodies of characters with the same form.

- The character hamza: Some letters have a zigzag form known as "Hamza". This form is considered a vowel in the Arabic alphabet.

- Vowels in Arabic script: Arabic vowels are placed above or below the body of characters. They are seven in number

## III. EXTRACTION STRUCTURAL CHARACTERISTICS

The extraction of structural features is based on three steps: pre-treatment, the determination of the baseline, and the detection of primitives.

The words to be recognized are extracted from their contexts check or postal letters. A stage analysis, segmentation and filtering of documents is required. This task is not part of our work. The words are supposed to be taken out of context without noise. Since our method of feature extraction is based mainly on the outline, the preprocessing step we have introduced in our system is the expansion in order to obtain a closed contour with the least points of intersections. Since the elimination of the slope may introduce additional distortions we have tried to avoid this step. It is for this reason that techniques of preprocessing, avoiding the inclination correction has emerged [4, 7, 18, 19].

### A. Determination of baselines

From the word we can extract two baselines. A upper and lower baseline. These two baselines divide the word into three regions. The poles "H" and diacritical dots high "P" which are regions above the upper baseline. The jambs "J" and diacritical dots lower "Q" correspond to regions below the lower baseline. The body of the word is the region between the two baselines. In general, the loops are in the middle. Figure 1 illustrates this subdivision.

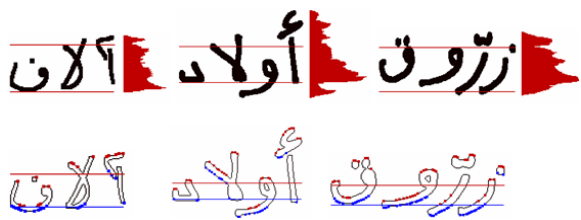

Figure 1. Improving the baselines by the method.

### B. Extraction of the poles and jambs

A pole is all forms with a maximum above the upper baseline. Similarly, jambs and all maxima below the lower baseline. The distance between these extrema and the baseline is determined empirically. It corresponds to:

*MargeH = 2(lower baseline – upper baseline) for the poles*

*MargeJ = (lower baseline – upper baseline) for the jambs.*

### C. Detection of diacritical dots

The diacritical points are extracted from the contour. Browsing through it, we can detect those that are closed. From these closed contours, we choose those with a number of contour points below a certain threshold. This threshold is derived from a statistical study to recognize the words taken from their context (checks, letters, mailing ...). It is estimated in our case the recognition of literal amounts to 60 pixels.

### D. Determination of loops from the contour

A closed contour length below 60 pixels corresponds to a loop if some contour points are between the two baselines. The problems encountered during the extraction of loops are:

- Some diacritical dots can be confused with the loops if they are intersecting with the baselines.
- Some loops may be longer than 60 pixels and can not be taken into account.

After a first selection step loops, a second step of verifying their inclusion in another closed loop is completed. This method involves:

- Looking for word parts that can include the loop.
- Stain the relevant section blank, if the contour points disappear when the latter is included in the word color and is part of the list of loops.

### E. Detection of PAWS

Given the variability of the shape of characters according to their position, an Arabic word can be composed by more than one party called for PAW "Pieces of Arabic Word." Detection of PAWS is useful information both in the recognition step in the step of determining the position of structural features in the word.

### F. Position detection primitives (letters)

The shape of an Arabic character depends on its position in the word. A character can have four different positions which depend on its position in the word. We can have single characters at the beginning, middle or end of a word. This position is detected during the primary feature extraction. Indeed, the extracted areas are defined by local minima. These minimums are from the vertical projection and contour. The number of black pixels is calculated in the vicinity of boundaries demarcated areas and between the two baselines above and below. If this number is greater than 0 at the left boundary and equal to 0 on the right, the position is the top "D", etc ... Figure 2 shows the various positions found in the Arabic script.

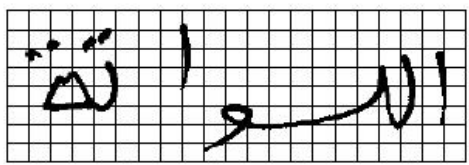

| Nb PAW | PAW1 | PAW2 | PAW3 | PAW4 |
|---|---|---|---|---|
| 4 | HI | HD HM BJF | HI | PD BPF |

Figure 2. Structural features for Arabic handwritten word.

## IV. DISCRIMINATION BETWEEN SCRIPTS

The recognition of the language of the document is regarded as a preprocessing step; this step has become difficult in the case of handwritten document. In this work we present a case of discrimination between Arabic script and Latin script by the structural method.

The preliminary study carried out, shows that the majority of the methods of differentiation treat only the printed text documents. Among the latter, the method suggested by [9], develops a strategy of discrimination between Arabic and Latin scripts. This approach is based on Template-Matching, which makes it possible to decide the identified language or the rejection. In [16], the authors uses a Multi-Classes supervised for classification and the Gabor filter to identify the Latin writing. The type of document used is printed with Latin writings and mixed. Two methods are proposed by [8], with two approaches: Statistical and spectral by Gabor filter. This work is tested on Kannada and Latin scripts. The system of identification, proposed by [11], relates to Latin and not-Latin languages in printed documents. This method is based on the application of Gabor filter. As well as the author classifiers for the identification of the languages other than Latin. With statistical methods and on printed documents, [22] has interested by identification of Arabic, Chinese, Latin, Devanagari, and Bangla languages. The identification of the type of scripts (printed or handwritten) is treated by [12], on Korean language. This approach is based on an analysis of related components and contours. A spectral method is presented in [23]. This method is to classes the script for Chinese, Japanese, Korean or Latin language by Gabor filter.

The Arabic script is cursive and present various diacritic. An Arab word is a sequence of named entirely disjoined related entities. Contrary to the Latin script, the Arab characters are written from right to left, and do not comprise capital letters. The characters form varies according to their position in the word: initial, median, final and isolated. In the case of handwritten, the characters, Arabic or Latin, can vary in their static and dynamic properties. The static variations relate to the size and the form, while the dynamic variations relate to the number of diacritic segments and their order.

Considering the visual difference between Arabic script and Latin script, we have chosen to study the possibility of discrimination between them based on the general structure of each, and the number of occurrences of the structural features mentioned above.

### A. Structural features of Arabic and Latin scripts

The Arabic alphabet has 28 letters have different shapes. These forms vary according to their positions in the word, their widths, number and position of the diacritical points. The Latin alphabet contains 26 letters also have different shapes. These forms vary according to their statements lowercase or uppercase the presence of accents in the character, items, and other forms that vary by language. These factors increase the total number of different types of characters to more than 100 for each of the scripts (Arabic and Latin). Figure 3 shows an example of different forms for Arabic letter and a Latin letter.

A first step is to see the structural aspect of each script of Arabic and Latin. We note though, that the general shape of the Arabic script is totally different from the Latin script, which is reflected in Table 1.

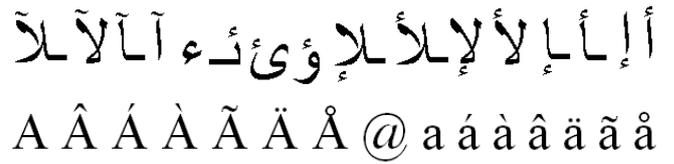

Figure 3.   Example of different forms for Arabic letter and Latin letter

Table 1 presents each structural feature used with their many appearances in letters, taking into account the different forms of each letter.

TABLE I.     PRESENCE OF STRUCTURAL FEATURES IN ARABIC AND LATIN SCRIPTS.

|                          | Arabic 120 forms | Latin 103 forms |
|--------------------------|------------------|-----------------|
| **H: ascender**          | 29               | 29              |
| **B: loop**              | 22               | 34              |
| **P: upper diacritic dots** | 30            | 28              |
| **J: descender**         | 28               | 12              |
| **Q: lower diacritic dots** | 11            | 0               |

The previous table shows that there is a difference between Arabic script and Latin script under these structural features. Latin script is characterized by:

- Many interesting number of loops.
- The interesting number of upper diacritics dots.
- The lack of lower diacritic dots.
- The small number of the descenders.

The Arabic script is characterized by:

- The average number of loops.
- The interesting number of upper diacritics dots.
- The interesting number of lower diacritic dots.
- The interesting number of the descenders.

### B. Performance of the method

This evaluation is performed on the *set_a* of IFN / ENIT database described in [24]. An estimate rate of the extraction compared to manual removal of files deducted *.tru* whose content is automatically generated from the printed script for each image of set_a, and manually verified by the crew of LSTS - ENIT. We present in Table 2 the first stage of evaluation. It is to count the number of each feature: Hampe (H), Jamb (J), Upper (P) and Lower (Q) diacritic dots, Loop (B), Start (D), Middle (M), End (F) and Isolated (I)

TABLE II.  EVALUATION RESULTS OF STRUCTURAL PRIMITIVE METHOD ON SET_A OF IFN / ENIT DATABASE.

| Feature | Total Set_a | Correctly extracted | Error rate |
|---|---|---|---|
| H | 16440 | 10852 | 33,99 % |
| J | 12632 | 8352 | 33,88 % |
| P | 12444 | 10165 | 18,31 % |
| Q | 7514 | 5957 | 20,72 % |
| B | 13149 | 9021 | 31,39 % |
| nbPAWs | 28312 | 1566 | 5.53% |

We note that the error rates are the highest for the poles and jambs. Given that these characteristics depend strongly on the quality of the baseline, a normalization step of words in order to eliminate the inclination and uniform size can improve the rate of extraction of these features. The best extraction rate obtained is that of PAWS despite confusion between this characteristic and diacritical dots may occur in the case of connection between diacritical points. To better understand the problem of error types, we conducted a second type of evaluation by analyzing the features extracted using their positions in the PAW.

In the second type of evaluation, we note first that all the characteristics Jamb is associated with the isolated position and end. The second point shows that the error rate is the highest in the extraction of Poles in the middle of PAW. The error rate for the other characteristics is more important when these characteristics are isolated. These two points are also mostly of poor extraction of the baseline. Indeed, in the handwriting, a pole is not in general long enough. Also seat-term impulses directly on the height of the shaft relative to the baseline above. For isolated letters, the baseline is not always correctly extracted, which explains the rate obtained.

*C. Evaluation*

The recognition of the language of the document is regarded as a preprocessing step, this step has become difficult in the case of handwritten document. We begin by discriminating between an Arabic text and a Latin text (Figure 4) printed by the structural method. Considering the visual difference between writing Arabic and Latin script, we have chosen to discriminate between them based on the general structure of each, and the number of occurrences of the structural characteristics mentioned above. Indeed, in analyzing a text in Arabic and Latin text we can distinguish a difference in the cursivity, the number of presence of diacritical dots and leg in the Arabic script. To printed Latin script, it is composed mainly of isolated letters.

The first step in the process regardless of the Arabic script from a text document is extracted lines and PAWS. The extraction of lines is done by determining the upper and lower limit using the horizontal projection. For each line, there are the PAWS using the method of vertical projection. Each PAW will be awarded by a system for extracting structural features.

Extraction of structural features for text, is performed by the method mentioned previously by introducing text PAWs successively. In figure 4, we present the results of structural characteristics for an Arabic text and a Latin text have the same number of rows and same size character fetches. The results in the figure show the average after analysis of a set of multiple images. Depending on the specific results, the distinction between Arab and Latin entries is possible using the structural method.

اغتنمت فرصة وجودي بهذا الملتقى للتعرف على مجالات أخرى للبحث العلمي في إطار التعرف على الأشكال و تحليل الصورة والقراءة الآلية للكتابة و في مجالات أخرى كالبحث وتحليل المعلومة. كما أن هذا الملتقى كان فرصة لي لإجراء بعض النقاشات العلمية مع بعض الباحثين داخل الملتقى وخارجه, حيث قمت ببعض النقاشات مع باحثين في مجالات تقترب من مجال أبحاثي كما أجريت لقاءات مع بعض الأخصائيين في مجال تشخيص و معالجة الإعاقة السمعية قصد التعرف على نتائج الأبحاث في هذا المجال ولقد أشاد بعض الأطباء الفرنسيين بتقدم هذا المجال في تونس

(a)

La première étape dans le processus de distinction de l'écriture arabe à partir d'un document texte est l'extraction des lignes, des PAWs et des lettres isolées. Afin d'éviter le problème de segmentation nous avons commencé par une discrimination entre les textes imprimés. L'extraction des lignes est effectuée par la détermination de la limite inférieure et supérieure en utilisant la projection horizontale.

(b)

Figure 4. Example of Arabic text (a) and Latin text (b) printed to discriminate.

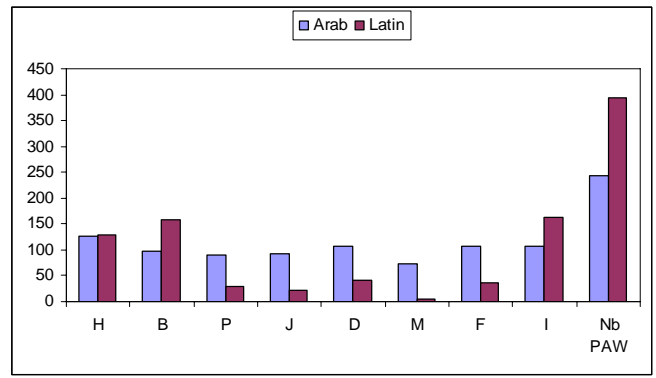

Figure 5. Results of the evaluation step of structural primitives.

V.  CONCLUSION

According to the study, the distinction between Arabic and Latin is possible, using the structural features. The results in Figure 5, show the different numbers of occurrence of each feature used for Arabic and Latin. These tests are performed on multiple images, and each document has the same impact. So, as there is a difference between Arab and Latin structures visually, the use of structural features shows that difference. Using a classifier is recommended in cases of discrimination, but since the distinction is fully visible, we have no interest to do so. Currently, tests are performed on documents written in Arabic or Latin. We expect shortly to continue to discriminate in the case of mixed paper (Arabic and Latin at the same document) and add other features that may improve the identification of language in critical cases, such as the presence of a Latin word in an Arabic text.